\PassOptionsToPackage{safe}{pbalance}
\documentclass[sigconf,nonacm]{acmart}
\usepackage{amsmath,booktabs,graphicx,array}
\setcopyright{none}
\setkeys{acmart.cls}{pbalance=true}
\acmDOI{}
\acmISBN{}
\title{Incremental Predictive Value of LLM Tutoring Interactions for Student Performance}

\author{Zeyu Dong}
\authornote{Corresponding authors.}
\email{dongzeyu123@outlook.com}
\affiliation{
  \institution{Crestwood Preparatory College}
  \city{Toronto}
  \state{Ontario}
  \country{Canada}
}

\author{Mikael Kwame Kuada}
\email{mikekuada12@gmail.com}
\affiliation{
  \institution{Hammond High Magnet School}
  \city{Hammond}
  \state{Louisiana}
  \country{United States}
}

\author{Asa Holley Buchakjian-Tweedy}
\email{asaholleybt@gmail.com}
\affiliation{
  \institution{Spaulding High School}
  \city{Rochester}
  \state{New Hampshire}
  \country{United States}
}

\author{Ibrahem Aljabea}
\authornotemark[1]
\email{ialjab2@lsu.edu}
\affiliation{
  \institution{Louisiana State University}
  \city{Baton Rouge}
  \state{Louisiana}
  \country{United States}
}
\renewcommand{\shortauthors}{Dong et al.}

\begin{document}
\begin{abstract}
We evaluate whether semester-level interaction features from a large language
model tutor improve assignment-score prediction beyond constant baselines.
StudyChat provides 16,851 student--assistant exchanges from two offerings of
an undergraduate artificial intelligence course, with outcomes matched to
175 of 203 student-semester records. Six fitted estimators and training-fold
mean and median baselines share repeated nested cross-validation. After
averaging each student's held-out predictions across ten repeats, Gradient
Boosting achieves mean absolute error of 3.023 grade points, versus 3.138
for the median baseline. The paired difference is $-0.114$, with a conditional
95\% bootstrap interval of $[-0.245,0.013]$. Representation changes and a
semester-specific median yield similarly small gains. In the lowest
observed-score quartile, however, boosting's error is 7.30 points, versus
8.07 for the median and 6.67 for the mean; Ridge has the lowest fitted-model
error at 6.37 despite its worse overall error of 3.331. Scores concentrate
near the ceiling: 160 students (91.4\%) score above 90, and the lowest quartile
contributes 62.1\% of boosting's total absolute error. This retrospective
comparison shows how baseline choice and error distribution affect predictive
value, informing evaluation of student-support tools without establishing a
learning effect.

\end{abstract}
\maketitle
\hypersetup{
  pdfauthor={Zeyu Dong, Mikael Kwame Kuada, Asa Holley Buchakjian-Tweedy, Ibrahem Aljabea},
  pdftitle={Incremental Predictive Value of LLM Tutoring Interactions for Student Performance},
  pdfkeywords={educational data mining, learning analytics, large language models, baseline evaluation, responsible AI}
}

\section{Introduction}
Conversational tutoring logs record how students seek help, offering a possible
basis for feedback and instructor support. Before such traces inform educational
decisions, their predictive value must be separated from the performance of a
simple baseline. When scores concentrate near the ceiling, a constant prediction
can achieve small average error while poorly representing lower-scoring students.
Evaluating only overall accuracy can therefore obscure the errors most relevant
to a support decision.

We examine this distinction in StudyChat~\cite{mcnichols2026studychat}, a
research release with 16,851 exchanges from two offerings of one undergraduate
artificial intelligence course. Semester-level features are available for
203 student-semester records, of which 175 have assignment outcomes. Among
these matched records, 160 (91.4\%) score above 90. The supervised sample
therefore consists of 175 outcome-bearing records, not 16,851 independent
learning outcomes.

StudyChat's original assessment-specific regressions add usage and dialogue-act
counts to prior assessment scores~\cite{mcnichols2026studychat}. Changes in
explained variance in that setting do not establish held-out prediction of a
semester aggregate or describe how prediction errors are distributed.
We address that evaluation gap through two questions: do semester-level
interaction summaries reduce held-out assignment-score error relative to
training-fold constants, and do baseline and estimator comparisons change
across observed score levels? This is a complementary analysis of the same
corpus, not a test of added value beyond prior academic performance.

The main finding is a contrast between small aggregate gains and uneven error.
Gradient Boosting's student-averaged mean absolute error (MAE) is 3.023 points,
versus 3.138 for a median baseline, with a conditional 95\% interval for the
difference crossing zero. In the lowest observed-score quartile, boosting
outperforms the median but not the mean baseline; Ridge has the smallest
fitted-model error there despite its worse overall rank.
The baseline and evaluation population both affect the apparent predictive value.

The analysis provides three connected contributions:
\begin{itemize}
\item A repeated nested comparison of six fitted estimators and two constant
baselines quantifies the incremental value of 14 interaction features.
\item All-model outcome-stratum contrasts identify where aggregate rankings
change and separate learner performance from baseline choice.
\item Representation, cohort-transfer, and held-out attribution checks examine
the scope of the predictive result and its interpretation.
\end{itemize}
For learning analytics, these results establish evaluation requirements before
predictions guide support; no support intervention is tested. Full-semester
features make this a retrospective study. Prospective evaluation requires
verified assessment-relative cutoffs.

\begin{figure*}[t]
\centering
\includegraphics[width=0.75\textwidth]{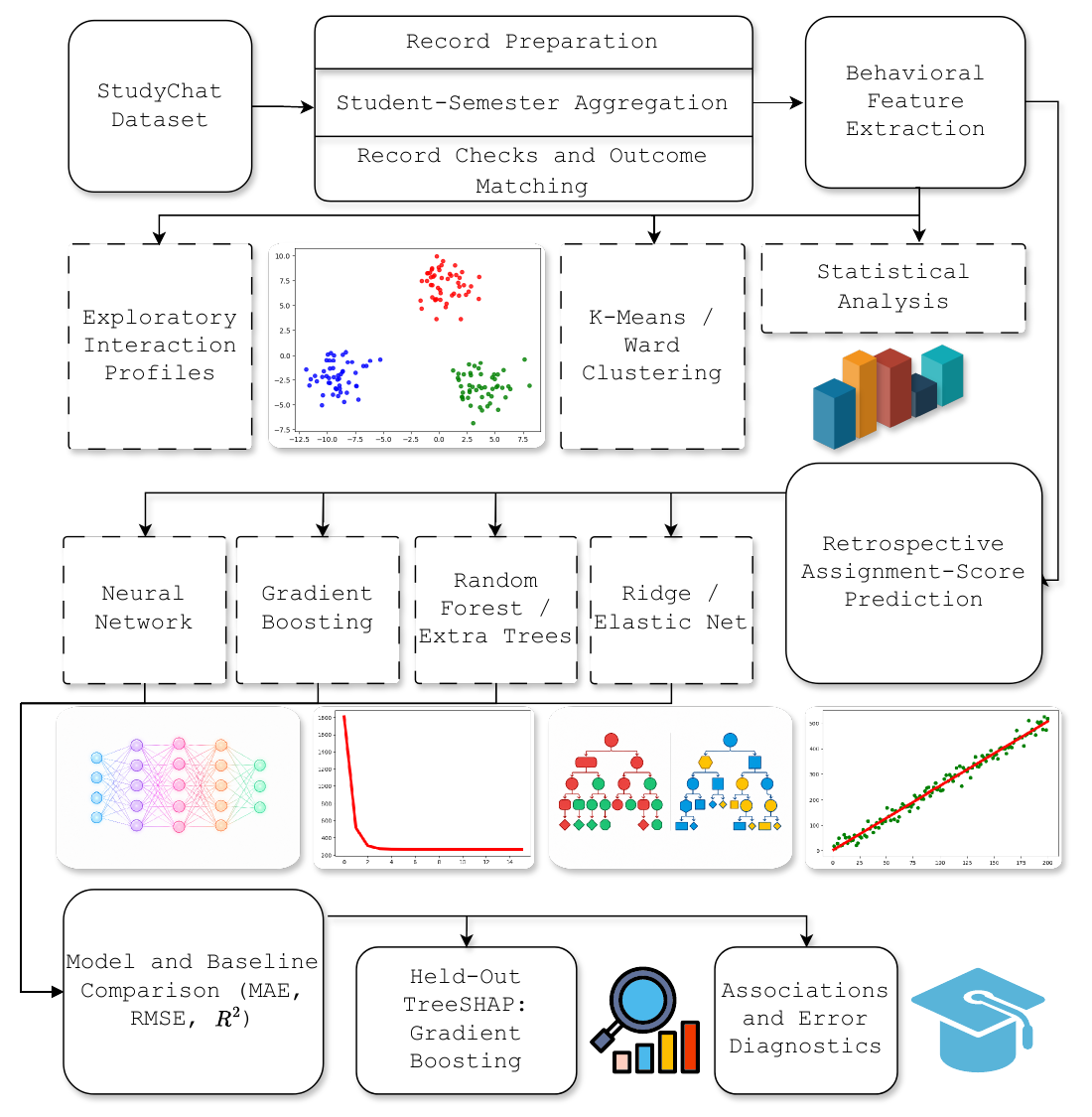}
\caption{Study workflow. Student-semester aggregation precedes repeated nested
prediction; imputation and applicable scaling are learned within training
partitions. Baseline contrasts assess predictive value. Clustering and
attributions are separate exploratory diagnostics, not validated learning
styles or intervention rules. Small inset plots are schematic, not study results.}
\Description{The pipeline separates data preparation, feature construction, retrospective prediction, exploratory clustering, and held-out model diagnostics.}
\label{fig:sac-pipeline}
\end{figure*}

Figure~\ref{fig:sac-pipeline} distinguishes the supervised evaluation from
the exploratory descriptions of interaction patterns.

\section{Related Work}
\paragraph{AI assistance and educational outcomes.}
Research on AI in education spans tutoring, feedback, assessment, and learner
support. Reviews describe technical adoption alongside uneven educator
involvement in evaluation~\cite{zawackirichter2019systematic}. Kasneci et al.\
discuss opportunities and risks of language-model assistance, including the
need for pedagogical oversight~\cite{kasneci2023chatgpt}. Bastani et al.\
compare tutor designs with and without learning safeguards in high-school
mathematics~\cite{bastani2025generative}. Lau and Guo interview programming
instructors about their planned responses to AI coding tools~\cite{lau2023from}.
These studies distinguish educational practice and learning effects from
the prediction of observed assignment scores addressed here.

\paragraph{Learning analytics and the evaluation gap.}
Learning analytics relates traces to educational processes and
outcomes~\cite{siemens2013learning,ferguson2012learning}, while educational data
mining uses submissions, assessments, and activity records for
prediction~\cite{romero2020educational}. StudyChat supplies dialogues, generated
dialogue-act labels, and normalized outcomes~\cite{mcnichols2026studychat}.
Its assessment-specific regressions add utterance or dialogue-act counts to
a prior-score baseline and examine explained variance and coefficients.
We instead compare held-out MAE for a semester-level aggregate on shared
nested splits and examine errors across observed outcomes. The different
target, information set, and error estimand distinguish the two analyses;
using the same corpus does not provide independent-cohort replication.

\paragraph{Prediction and explanation.}
Nested cross-validation separates parameter search from outer
evaluation~\cite{varma2006bias,cawley2010overfitting}. Repeated folds remain
dependent, so we distinguish descriptive fold averages from fixed-fit
student-bootstrap contrasts~\cite{efron1979bootstrap}. SHAP decomposes a fitted
prediction into additive contributions~\cite{lundberg2017unified}; numerical
additivity alone does not establish accuracy or educational validity.
These distinctions matter when explanations might otherwise be read as
recommendations about how a student should learn.

\section{Data and Evaluation}
\subsection{Cohort, Target, and Information Available}
StudyChat covers COMPSCI 383 at the University of Massachusetts Amherst in
Fall 2024 and Spring 2025~\cite{mcnichols2026studychat}. Joining on student and
semester matches 65 of 84 Fall records and 110 of 119 Spring records to
nonmissing outcomes. Of 181 grade records, six lack matching behavioral
records; 28 behavioral records lack outcomes. Missing outcomes are not imputed.

The target is $y_i=100\,\texttt{llm}_i$, the released points-weighted aggregate
of assignments a1--a7, rather than an exam or overall course grade.
Assignment maxima are $(30,100,55,61,65,113,101)$ in Fall and
$(31,100,55,90,105,113,135)$ in Spring, with nominal totals 525 and 629.
The inspected release metadata does not document missing/unsubmitted-item
treatment. We inherit the nonmissing aggregate rather than reconstructing it.
Matched scores have mean 94.77, median 96.18, and sample standard deviation
6.12; six are below 80. The pooled evaluation gives each record equal weight,
with 65/175 from Fall and 110/175 from Spring.

Preparation parses millisecond timestamps, aggregates student-semester records,
and checks unique keys and joins; raw-exchange deduplication is not performed.
The primary representation contains 14 operational predictors
(Table~\ref{tab:sac-features}), excluding conversation persistence ($P/C-1$)
from the original 15 features, where $P$ counts exchanges and $C$ counts chats.
Identifiers and semester are not learned-model
predictors. Contextual-question and writing-request ratios use GPT-4.1-generated
dialogue-act annotations, stored as \path{clarification_ratio} and
\path{solution_request_ratio}. These are recorded language categories, not
validated measures of motivation, metacognition, or learning styles.

\begin{table}[t]
\caption{Primary features. $P$: exchanges; $C$: chats; $W$: prompt tokens;
$U$: distinct prompt tokens. Tokens match lowercase \texttt{[a-z0-9']+}.
Percentages use the 0--100 scale.}
\label{tab:sac-features}
\centering
\begin{tabular}{>{\raggedright\arraybackslash}p{0.32\columnwidth}>{\raggedright\arraybackslash}p{0.59\columnwidth}}
\toprule
Feature & Operational definition \\
\midrule
Prompt count & $P$ exchanges. \\
Conversation count & $C$ distinct threads. \\
Dialogue depth & Mean maximum non-system history count per chat. \\
Prompt length & $W/P$ regex tokens. \\
Vocabulary diversity & $U/W$. \\
Follow-up ratio & Percentage with \texttt{interactionCount}>0. \\
Contextual-question ratio & Percentage labeled contextual questions. \\
Example-pattern ratio & Percentage matching example-request text. \\
Writing-request ratio & Percentage labeled writing requests. \\
Conversation span & Mean last-minus-first chat timestamp in minutes, including idle time. \\
Response length & Response tokens divided by $P$. \\
Questions per chat & Prompts containing a question mark divided by $C$, with at most one count per prompt. \\
Tokens per segment & Tokens per segment split on \texttt{[.!?\textbackslash n]+}; at least one segment per prompt. \\
Active days & Distinct UTC dates. \\
\bottomrule
\end{tabular}
\end{table}

The example pattern matches ``examples'', ``e.g.'', ``for instance'',
``show me'', ``demonstrate'', ``demo'', or ``sample''.
Long legacy chat spans and one large tokens-per-segment value
are retained and examined through sensitivity analyses. All features summarize
the full observed semester. Outcomes do not enter feature construction, but
activity may occur during or after target assignments. Official deadlines
and grade-availability times are unavailable for a verified forecast cutoff.

\subsection{Models and Nested Evaluation}
Mean and median dummy regressors are compared with Ridge, Elastic Net,
Random Forest~\cite{breiman2001random}, Extra Trees~\cite{geurts2006extremely},
Gradient Boosting~\cite{friedman2001greedy}, and a Keras multilayer perceptron
(MLP). Non-neural models use scikit-learn~\cite{pedregosa2011scikit}.
All eight estimators share five outer folds repeated ten times (seed 42).
Each outer-training partition has four shuffled inner folds, selecting
parameters by validation MAE with seed $42+1000r+f$ for zero-based repeat
$r$ and fold $f$. Search uses 40 randomized candidates where applicable,
all 25 Ridge values, and eight neural candidates. Each dummy predicts the
mean or median outcome of its outer-training partition; held-out outcomes
do not determine those predictions.

Each inner-training fit and outer refit learns median imputation from its
training partition. Ridge, Elastic Net, and MLP also apply $\log(1+x)$ to
prompt/conversation counts, depth, prompt/response lengths, span,
questions per chat, tokens per segment, and active days, then fit
standardization. Trees retain unlogged features.

\begin{table*}[t]
\caption{Available search specification. Random Forest and Extra Trees use
500 trees. Resolved budgets and selected parameters are saved; the full
historical sampled-candidate trace and dirty source are not archived.
Boosting's search definition is available in later versioned source and
agrees with selected parameters, not a certified original source snapshot.}
\label{tab:sac-search}
\centering
\begin{tabular}{p{0.14\textwidth}>{\raggedright\arraybackslash}p{0.79\textwidth}}
\toprule
Estimator & Candidates \\
\midrule
Ridge & 25 log-spaced $\alpha$ values in $[10^{-4},10^4]$. \\
Elastic Net & 20 log-spaced $\alpha$ values in $[10^{-4},10]$; $L_1$ ratio $\{0.1,0.25,0.5,0.75,0.9,1\}$. \\
Random Forest & Squared/absolute-error criterion; depth $\{2,3,4,5,\mathrm{unlimited}\}$; minimum split $\{4,8,12,20\}$; leaf $\{2,3,5,8,12\}$; feature sampling \texttt{sqrt} or $\{0.5,0.75,1\}$; sample fraction $\{0.6,0.8,1\}$. \\
Extra Trees & Squared error; same depth, split, leaf, and feature choices as Random Forest; no sample-fraction search. \\
Gradient Boosting & Squared/Huber/absolute-error loss; trees $\{50,100,200,300\}$; rate $\{0.01,0.03,0.05,0.1\}$; depth $\{1,2,3\}$; leaf $\{3,5,8,12\}$; subsample $\{0.6,0.8,1\}$; features \texttt{sqrt} or $\{0.5,1\}$. \\
MLP & Layers $(8)$, $(16,8)$, $(32,16)$; dropout $\{0,0.1,0.2\}$; $L_2$ $\{10^{-4},10^{-3},10^{-2}\}$; rate $\{3\times10^{-4},10^{-3},3\times10^{-3}\}$. \\
\bottomrule
\end{tabular}
\end{table*}

In Table~\ref{tab:sac-search}, numeric feature-sampling choices are fractions
of the 14 predictors; \texttt{sqrt} selects their square-root count.

The historical MLP uses a PyTorch backend, ReLU hidden layers, linear output,
dropout after the first hidden layer, Huber loss ($\delta=3$), Adam, batch
size 16, and at most 500 epochs. Early stopping uses an internal 20\% split,
patience 35, minimum improvement $10^{-4}$, and restored best weights.
The learning rate halves after 12 stagnant epochs, down to $10^{-5}$.
Search uses one seed; outer refits average seeds 42, 43, and 44.
Preprocessing is fitted before the internal early-stopping split, so those
validation rows participate in fitted transformations; outer held-out rows do not.

\subsection{Estimands and Uncertainty}
For $n=175$ matched records, let $\hat y_{ir}$ be record $i$'s held-out
prediction in repeat $r$. We report the error after averaging its $R=10$
predictions:
\begin{equation}
\bar{\hat y}_i=R^{-1}\sum_r\hat y_{ir},\qquad
\mathrm{MAE}_{\mathrm{student}}=n^{-1}\sum_i|y_i-\bar{\hat y}_i|.
\end{equation}
This repeated-cross-validation aggregate differs from a single deployment fit.
We also report mean fold MAE, root mean squared error (RMSE), and $R^2$.
For training-fold constant baseline $b$, the paired contrast is
\begin{equation}
\Delta_b=n^{-1}\sum_i\bigl(|y_i-\bar{\hat y}_{i,\mathrm{model}}|
-|y_i-\bar{\hat y}_{i,b}|\bigr).
\end{equation}
Negative differences favor the learner. The median is the lower-error constant
under both primary MAE summaries. Paired 95\% percentile intervals use 5,000
student resamples; marginal MAE intervals use 2,000. Predictions stay fixed:
intervals do not include retraining, tuning, comparator/model selection, or
multiplicity adjustments. The 0.5-point reduction reference is an operational
reporting convention, not a preregistered or validated educational threshold.

Post-hoc outcome quartiles use \texttt{pandas.qcut}, right-closed with the
minimum included and ties kept together. Rounded cutpoints are 94.286,
96.184, and 97.905; counts are 45, 44, 44, and 42.
Within-stratum intervals use 5,000 paired resamples (seed 42), conditional on
the same averaged predictions and unadjusted across models/strata.

\section{Results}
\subsection{Incremental Prediction}
Table~\ref{tab:sac-models} reports the complete primary comparison.
MLP has the smallest mean fold MAE (3.061), followed by boosting (3.062)
and Random Forest (3.071). Their student-averaged MAEs are 3.027, 3.023,
and 3.038, versus 3.138 for the median. Ridge, Elastic Net, and Extra Trees
have larger point errors; all six fitted models have negative mean fold $R^2$.

\begin{table*}[t]
\caption{All eight estimators on 50 shared folds and 175 matched records.
Student MAE scores ten-repeat-averaged predictions. Paired differences use
the median dummy and conditional 95\% percentile intervals. All six
learner intervals include zero. All displayed metrics are rounded once
from their underlying values to three decimals.}
\label{tab:sac-models}
\centering
\begin{tabular}{lrrrrr}
\toprule
Model & Fold MAE & Fold RMSE & Fold $R^2$ & Student MAE & Paired $\Delta$ [95\% interval] \\
\midrule
Mean dummy & 3.446 & 5.713 & $-0.147$ & 3.440 & $+0.303\ [+0.113, +0.485]$ \\
Median dummy & 3.141 & 5.758 & $-0.082$ & 3.138 & --- \\
Ridge & 3.356 & 5.578 & $-0.113$ & 3.331 & $+0.193\ [-0.052, +0.442]$ \\
Elastic Net & 3.383 & 5.601 & $-0.123$ & 3.344 & $+0.206\ [-0.022, +0.440]$ \\
Random Forest & 3.071 & 5.523 & $-0.023$ & 3.038 & $-0.100\ [-0.243, +0.043]$ \\
Extra Trees & 3.310 & 5.547 & $-0.106$ & 3.282 & $+0.144\ [-0.083, +0.374]$ \\
Gradient Boosting & 3.062 & 5.528 & $-0.012$ & 3.023 & $-0.114\ [-0.245, +0.013]$ \\
Keras MLP & 3.061 & 5.479 & $-0.004$ & 3.027 & $-0.111\ [-0.255, +0.029]$ \\
\bottomrule
\end{tabular}

\end{table*}

Boosting's point reduction in student MAE is 0.1144 points. Its signed
model-minus-median difference has conditional 95\% interval
$[-0.2451,0.0125]$. MLP and Random Forest have point reductions of
0.1107 and 0.0998 points, respectively; their intervals are
$[-0.2546,0.0290]$ and $[-0.2434,0.0428]$. All three intervals include
zero, and the point gains remain below 0.5 points.

The recorded diagnostic-model rule first retains learners that beat the
best dummy in fold MAE, then favors a lower-complexity family within
0.5 points of the lowest error. Boosting and Random Forest share the
tree-family rank; boosting's lower error breaks that tie, while MLP ranks
as more complex. This outer-result-dependent choice makes subsequent
selected-model diagnostics exploratory. The predictive claim rests on all models.

\subsection{Baseline Dependence and Error Distribution}
Boosting's MAEs across Q1--Q4 are 7.30, 0.91, 1.04, and 2.74.
The 45 Q1 students contribute 62.1\% of its total absolute error.
In Q1, boosting's difference from the median is $-0.767$
$[-1.091,-0.457]$, but the difference from the mean is $+0.629$
$[0.294,0.944]$. The mean's lower prediction level is advantageous
there despite its worse overall MAE of 3.440.

\begin{table}[t]
\caption{Student-averaged MAE by observed-score quartile for all estimators.
Counts are 45, 44, 44, and 42; their weighted errors reproduce overall MAE.}
\label{tab:sac-tail}
\centering
\setlength{\tabcolsep}{2.5pt}
\begin{tabular}{lrrrrr}
\toprule
Model & Overall MAE & Q1 & Q2 & Q3 & Q4 \\
\midrule
Mean dummy & 3.440 & 6.669 & 0.673 & 2.405 & 3.965 \\
Median dummy & 3.138 & 8.065 & 0.799 & 0.996 & 2.551 \\
Ridge & 3.331 & 6.368 & 0.879 & 1.936 & 4.107 \\
Elastic Net & 3.344 & 6.445 & 0.771 & 2.015 & 4.108 \\
Random Forest & 3.038 & 7.282 & 0.832 & 1.035 & 2.899 \\
Extra Trees & 3.282 & 6.418 & 0.762 & 1.960 & 3.946 \\
Gradient Boosting & 3.023 & 7.298 & 0.909 & 1.036 & 2.739 \\
Keras MLP & 3.027 & 7.031 & 0.832 & 1.210 & 2.940 \\
\bottomrule
\end{tabular}

\end{table}

\begin{figure*}[t]
\centering
\includegraphics[width=0.90\textwidth]{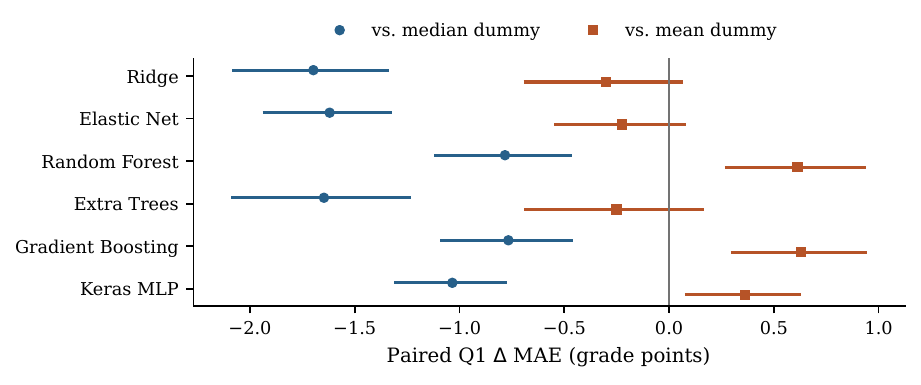}
\caption{Paired Q1 contrasts for all six learners on the same 45 records.
Circles compare with the median; squares compare with the mean. Lines are
conditional 95\% percentile intervals from 5,000 paired within-stratum draws.
Negative differences favor the learner. Strata are defined after observing
outcomes; intervals are not multiplicity-adjusted.}
\Description{All model-minus-median intervals are negative. Against the mean, three intervals cross zero and three are positive.}
\label{fig:sac-q1}
\end{figure*}

Ridge has the smallest fitted-model Q1 error, 6.37, but its mean-relative
difference $-0.301$ $[-0.694,0.066]$ includes zero.
All six learners' Q1 intervals lie below zero against the median;
none lies wholly below zero against the mean
(Figure~\ref{fig:sac-q1}). Apparent improvement against one constant
therefore does not alone establish added lower-tail information.
Q1 is the lowest quarter of a high-scoring cohort, not a prospective risk group.

For quartile $q$ with $n_q$ students, let $\mathrm{MAE}_q$ denote its
mean absolute error and $c_q$ its share of total absolute error. Then
\begin{equation}
\begin{aligned}
\mathrm{MAE}_{\mathrm{student}}
&=\sum_{q=1}^{4}\frac{n_q}{n}\mathrm{MAE}_q,\\
c_q&=\frac{n_q\mathrm{MAE}_q}{n\,\mathrm{MAE}_{\mathrm{student}}}.
\end{aligned}
\label{eq:quartile-aggregation}
\end{equation}
Thus, Q1 contains 25.7\% of the matched records but contributes 62.1\% of
boosting's absolute error. Improving a model's error in one stratum can be
offset by larger errors elsewhere; overall MAE alone does not show that trade-off.

\subsection{Representation and Cohort Sensitivity}
Boosting student MAE is 3.024 with all 15 features and 3.021 with robust
duration/segmentation definitions. The latter use a 30-minute inactivity
threshold, a 60-minute within-session gap cap, and code stripping in
sentence segmentation while retaining the token numerator.
Removing the three request-type ratios gives
3.070; removing prompt/conversation counts and active days gives 3.079;
removing follow-up ratio gives 3.023. All five use 50 folds, and all
intervals against their corresponding median dummies include zero.
A training-fold median for each semester gives student MAE 3.134 and fold
MAE 3.138. Boosting's student-MAE difference is $-0.111$
$[-0.243,0.021]$ (5,000 paired draws, seed 42).

Transfer fits preprocessing, four-fold inner tuning, and the final estimator
on the source semester only. It retains candidate budgets, direction seeds
50042/50043, and one neural search seed with three final members.
Fall-to-Spring trains on 65 and tests on 110 records; reverse transfer
exchanges those counts (Table~\ref{tab:sac-transfer}). Boosting's gains over
the better constant are 0.113 and 0.031 points. Paired uncertainty is
unavailable because transfer prediction rows were not archived.

\begin{table}[t]
\caption{Transfer MAE within the same course. The better constant is the mean
for Fall-to-Spring and median for Spring-to-Fall. These are point estimates.}
\label{tab:sac-transfer}
\centering
\begin{tabular}{lrr}
\toprule
Estimator & Fall to Spring & Spring to Fall \\
\midrule
Mean dummy & 3.385 & 3.356 \\
Median dummy & 3.413 & 2.775 \\
Ridge & 3.331 & 3.046 \\
Elastic Net & 3.377 & 3.077 \\
Random Forest & 3.293 & 2.895 \\
Extra Trees & 3.346 & 3.101 \\
Gradient Boosting & 3.272 & 2.744 \\
Keras MLP & 3.342 & 2.879 \\
\bottomrule
\end{tabular}
\end{table}

Alternate-target runs retain the 175 students, 14 legacy predictors,
50 outer folds, and four-fold inner MAE selection with 40 candidates where
applicable. For the exam aggregate, the median MAE is 6.946 versus mean
7.098; Random Forest, Ridge, and boosting give 6.909, 6.939, and 6.958.
Their paired median-relative intervals all include zero:
$[-0.371,0.304]$, $[-0.323,0.309]$, and $[-0.251,0.288]$.
For the third exam, median MAE is 8.287 versus mean 8.612; boosting,
Random Forest, and Ridge give 8.387, 8.467, and 8.625, with intervals
$[-0.024,0.225]$, $[-0.025,0.387]$, and $[-0.063,0.733]$.
These are full-semester retrospective checks, not prospective exam forecasts.

\subsection{Associations and Explanation Fidelity}
Four of 14 Pearson tests pass Benjamini--Hochberg
$q<0.05$~\cite{benjamini1995controlling}: contextual-question ratio
($r=-0.329$), active days ($0.248$), vocabulary diversity ($-0.199$),
and conversation count ($0.197$). Adjustment covers that screen only.
Removing six scores below 80 changes the contextual-question correlation to
$-0.173$; its Spearman coefficient is $-0.162$. Thus, a nominally prominent
association is sensitive to the lower-score tail. It agrees in direction
with StudyChat's Assignment-5 result on the same corpus~\cite{mcnichols2026studychat}.
Dialogue depth and follow-up ratio also have absolute Spearman correlation 0.993.

\begin{table*}[t]
\caption{The four associations retained by the 14-test Pearson screen on
175 records. Adjusted $q$ values apply only to that screen. Pearson intervals
use 2,000 student resamples; Spearman coefficients provide a complementary
rank-based summary. Vocabulary diversity's interval includes zero despite
its adjusted parametric test, illustrating the different assumptions.}
\label{tab:sac-correlations}
\centering
\begin{tabular}{lrrrr}
\toprule
Feature & Pearson $r$ & Spearman $\rho$ & Adjusted $q$ & Interval for $r$ \\
\midrule
Contextual-question ratio & $-0.329$ & $-0.162$ & 0.0001 & $[-0.534,-0.106]$ \\
Active calendar days & 0.248 & 0.220 & 0.0067 & $[0.148,0.335]$ \\
Vocabulary diversity & $-0.199$ & $-0.082$ & 0.0315 & $[-0.385,0.003]$ \\
Conversation count & 0.197 & 0.250 & 0.0315 & $[0.101,0.282]$ \\
\bottomrule
\end{tabular}

\end{table*}
Table~\ref{tab:sac-correlations} reports both coefficient types and uncertainty.

Tree-path-dependent TreeSHAP explains transformed held-out boosting rows
as $f_m(\mathbf{x}_i)=\phi_{0m}+\sum_{j=1}^{14}\phi_{ijm}$.
Averaging each student's ten baseline/contribution vectors preserves the
decomposition of the averaged prediction. Refit predictions match saved
values within $1.42\times10^{-14}$ points; the maximum additivity residual
is $5.68\times10^{-14}$. These are numerical checks on evaluated inputs,
not certification of the historical source or educational validity.
Mean absolute contributions lead with contextual questions (0.332 points),
conversation count (0.258), and questions per chat (0.150). Yet median
absolute outcome error is 1.69 and the maximum is 40.93.
Precise explanation of a fitted prediction does not imply an accurate prediction,
and contribution magnitudes do not identify causal learning mechanisms.

\begin{figure*}[t]
\centering
\includegraphics[width=0.75\textwidth]{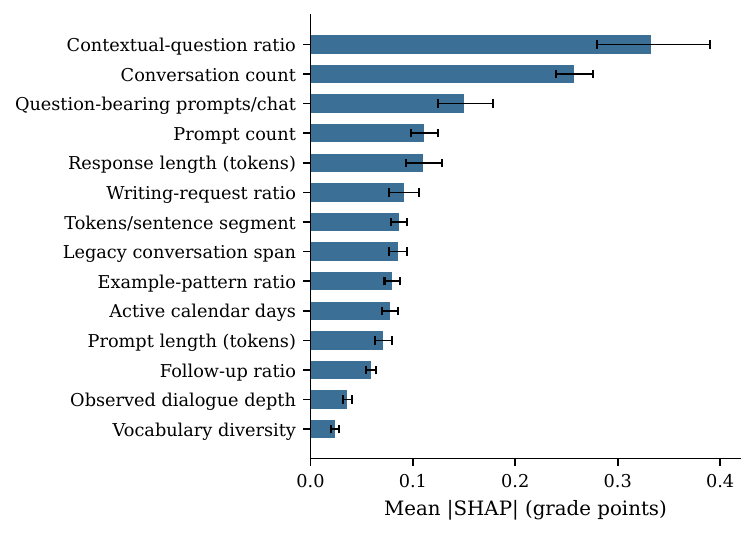}
\caption{Held-out attribution magnitudes for 14 predictors. Each bar is the
mean absolute ten-repeat-averaged contribution over 175 records; conditional
95\% percentile intervals use 2,000 student resamples. The ordering describes the selected boosting
function under tree-path-dependent TreeSHAP and correlated inputs.}
\Description{Contextual questions, conversation count, and questions per chat have the largest absolute SHAP contributions.}
\label{fig:sac-attributions}
\end{figure*}
Figure~\ref{fig:sac-attributions} shows all predictors rather than only the
largest contributions.

\subsection{Exploratory Interaction Profiles}
The companion clustering analysis uses eight standardized dialogue-act
percentages for all 203 records, separately from supervised predictors.
K-means~\cite{lloyd1982least} and Ward clustering describe request mixtures.
For $k=2,\ldots,6$, silhouette~\cite{rousseeuw1987silhouettes} selects six;
extending to twelve selects eight, while Davies--Bouldin~\cite{davies1979cluster}
selects ten. All evaluated partitions above six contain singletons.

Subset sensitivity samples 203 indices with replacement 200 times, discards
repetitions, and refits on distinct rows with fixed scaling. All original rows
are assigned to the nearest refitted centroid; adjusted Rand index
(ARI)~\cite{hubert1985comparing} compares assignments with the full-cohort
reference. Reference/subset fits use 20 initializations, versus 100 in
selection. This is not a multiplicity-preserving bootstrap.
At the operational median-ARI cutoff 0.60, $k=2,9,10$ pass. The finer two
contain singletons; $k=2$ has a wide central 90\% range, $[0.235,1.000]$.

The proportions are compositional~\cite{aitchison1982statistical,pal2022clustering}.
A separate gap analysis~\cite{tibshirani2001estimating} selects two by the
one-standard-error rule, but its uniform-reference draws do not preserve the
sum constraint. For the original six groups, Welch ($p=0.158$) and
Kruskal--Wallis ($p=0.138$) leave location evidence inconclusive; median-centered
Levene ($p=0.002783$) indicates unequal dispersion. These results do not
establish stable learning styles or validate adaptive instruction by cluster.

\section{Educational Implications and Scope}
\subsection{From Prediction to Student Support}
The study supplies an empirical check before interaction traces are used
to guide student support. The observed small gains provide little evidence
for allocating feedback or instructor attention from this representation
alone. Meanwhile, Q1 error concentration shows why population-average error
is insufficient for evaluating a tool intended to assist lower-performing
students. A support system should be assessed for its intended decision and
population, not adopted solely because its average score error is small.

For responsible use, educators should define the intended decision, the
information available when it is made, and the costs of missed support and
unnecessary alerts before fitting a model. Those choices determine which
baselines and evaluation populations are relevant. Our result concerns
prediction beyond constants; prior academic performance remains an important
untested comparator. It also does not imply that asking contextual questions
harms learning. Such questions may express confusion, task requirements,
productive help seeking, or annotation error.

Interaction measurement is distinct from learning-process measurement.
Counts and ratios discard
dialogue order, task semantics, response correctness, and verified timing.
The study does not directly measure motivation, emotion, or self-regulation.
An educator-reviewed application could investigate these dimensions, but the
present features should not be renamed as psychological constructs.
Likewise, outcome-defined error heterogeneity is not a demographic fairness audit.
Absence of protected-group evaluation precludes a claim of bias-free predictions.

\subsection{Selection, Ethics, and Reproducibility}
Outcome matching is 77.4\% in Fall and 92.4\% in Spring.
A saved exploratory comparison finds longer prompts among matched records:
standardized mean difference 0.463, with 95\% percentile interval
$[0.231,0.672]$. It uses 1,000 separate-group draws and standardizer
$\sqrt{(s_M^2+s_U^2)/2}$ from matched ($M$) and unmatched ($U$)
group sample variances. A shared PCG64
generator seeded at 0 preserves row order, draws matched before unmatched
samples, and processes prompt count, conversation count, depth, then prompt
length without resetting. The matched cohort differs from the behavioral
sample; unavailable outcomes leave the direction of selection bias unresolved.

Both offerings are from one course. Fixed-fit uncertainty excludes
retraining/search variation, and intervals crossing zero are not an equivalence
test. Generated act labels lack a supplied human construct-validation study.
Prior knowledge and outside study or AI use are unmeasured.

This secondary analysis uses the opt-in, de-identified research release.
Source authors report consent~\cite{mcnichols2026studychat}; no participants
were newly recruited, contacted, or assigned an intervention. We do not
claim a new institutional approval. Reporting uses aggregates rather than
participant text, identifiers, or row-level outcomes.

Saved predictions and resolved configuration support numerical replay of
the complete eight-estimator run. The original dirty training source was not
archived, and the recorded commit lacks boosting. A completion manifest
reports Keras 3.15.1, whereas a separate environment snapshot records it as
unavailable. Later versioned code documents search design, not exact historical
implementation. These limitations are distinct from numerical agreement with
the saved outputs. Replay checks all 400 model-fold target/semester orders;
raw fold values are averaged before rounding.

A separate source-frozen check retained the same cohort, splits, and search
budgets. In 90 minutes, both constants, Ridge, Elastic Net, and Random Forest
completed 50 folds each and matched point errors and paired intervals to four
decimals. Extra Trees was interrupted; boosting and cohort MLP were not reached.
This partial same-cohort check does not replace the historical comparison or
provide independent validation.

\subsection{Future Work}
Future evaluation should first fix prediction times before assignment deadlines
and restrict features to information available by each cutoff. Comparing
interaction summaries with prior assessment and course-context predictors on
held-out courses would test added value beyond existing academic records.
Richer representations could retain dialogue sequence, task content, and
response correctness, accompanied by human validation of dialogue-act labels
and evaluation of missing outcomes and retraining variation.

Educators and learners should then help define actionable support and acceptable
error costs. A controlled study of instructor-reviewed support could assess
whether earlier feedback improves subsequent unaided performance, retention,
or self-regulated learning. Prespecified measures of missed support,
unnecessary alerts, and instructor workload would distinguish predictive
improvements from educational benefit. None of these intervention outcomes
is established by the present retrospective evaluation.

\section{Conclusion}
Semester-level StudyChat features yield small overall assignment-score gains
over training-fold medians, while baseline and estimator rankings vary across
observed outcomes. Boosting's overall gain is 0.114 points with a conditional
interval crossing zero; its Q1 advantage over the median reverses against the
mean. All-model reporting and explicit error populations are therefore
necessary to interpret incremental predictive value in this cohort.
For learning analytics, the contribution is a baseline-referenced assessment
of this representation and the students who carry its error. It informs the
design and evaluation of support tools without demonstrating a learning effect.
Prospective cross-course prediction and an educator-informed intervention study
are the next tests of practical educational value.

\bibliographystyle{ACM-Reference-Format}
\bibliography{reference}
\end{document}